\documentclass{article}
\usepackage{spconf,amsmath,graphicx}

\usepackage[latin1]{inputenc}

\usepackage{amsfonts}
\usepackage{amssymb}
\usepackage{scalerel}

\usepackage[T1]{fontenc}

\usepackage{algorithmicx}
\usepackage{algpseudocode}
\usepackage{algorithm}

\usepackage{url}

\algrenewcommand\algorithmicindent{1.0em}%

\newcommand\scale[2]{\vstretch{#1}{\hstretch{#1}{#2}}}
\newcommand{\LIPplus}{\mathbin{\ooalign{$\bigtriangleup$\crcr\hidewidth
  \raise.14em\hbox{$\scale{0.7}{\scriptscriptstyle+}$}\hidewidth}}}
\newcommand{\LIPminus}{\mathbin{\ooalign{$\bigtriangleup$\crcr\hidewidth
  \raise.14em\hbox{$\scale{0.7}{\scriptscriptstyle-}$}\hidewidth}}}
\newcommand{\LIPtimes}{\mathbin{  \ooalign{$\bigtriangleup$\crcr\hidewidth
  \raise.14em\hbox{$\scale{0.7}{\scriptscriptstyle\times}$}\hidewidth}}}

\newcommand{\Real}{\mathbb R}
\newcommand{\Zint}{\mathbb Z}

\title{Speeding up the K\"ohler's method of contrast thresholding}
\name{Guillaume Noyel, Member, IEEE}
\address{International Prevention Research Institute\\
	95 cours Lafayette\\
	69006 Lyon, France}
\begin{document}
%\ninept
%
\maketitle
\begin{abstract}
K\"ohler's method is a useful multi-thresholding technique based on boundary contrast. However, the direct algorithm has a too high complexity - $\mathcal{O} (N^2)$ i.e. quadratic with the pixel numbers $N$ - to process images at a sufficient speed for practical applications. In this paper, a new algorithm to speed up K\"ohler's method is introduced with a complexity in $\mathcal{O}(NM)$, $M$ is the number of grey levels. The proposed algorithm is designed for parallelisation and vector processing, which are available in current processors, using OpenMP (Open Multi-Processing) and SIMD instructions (Single Instruction on Multiple Data). A fast implementation allows a gain factor of 405 in an image of 18 million pixels and a video processing in real time (gain factor of 96).
\end{abstract}
\begin{keywords}
K\"ohler multi-thresholding, boundary contrast, fast image segmentation, parallelisation, pattern recognition
\end{keywords}

%%%%%%%%%%%%%%%%%%%%%%%%%%%%%%%%%%%%%%%%%%%%%%%%%%%%%%%%%%%%%%%
%
%		Introduction
%
%%%%%%%%%%%%%%%%%%%%%%%%%%%%%%%%%%%%%%%%%%%%%%%%%%%%%%%%%%%%%%%
\section{Introduction}
\label{sec:intro}

Adaptive thresholding is one of the most used technique in many applications because it is fast to compute and when combined with previous filters, it gives robust decision rules for pattern recognition. Among many other techniques of thresholding \cite{Cai2014}; K\"ohler's method computes a curve of contrast of the region boundaries in an image \cite{Kohler1981}. The contrast steps correspond to the local maxima of the curve and can be extracted for (multi-)thresholding of the image. This is useful for many applications: industrial, biomedical, video, etc \cite{Hautiere2006,Jourlin2012,Jourlin2016_chap3}. However, computing K\"ohler's method is time consuming; almost 1 minute using a C++ implementation on a current computer with an image acquired by a recent camera (18 million pixels, fig. \ref{fig:pre:Kolher_tulips}). The purpose of this paper, is to introduce and implement a new algorithm for K\"ohler's thresholding method faster than the existing algorithms and making it useful for applications requiring fast or real-time processing (e.g. video thresholding, large datasets) \cite{Thomee2016}.

Previously, two attempts were made to speed up the computation of K\"olher's method. Zeboudj \cite{Zeboudj1988,Coster1989} used mathematical morphology operations to give a similar version of K\"ohler's method. However, his approach was efficient on specific devices which are not available any more. The other one, from Hauti\`ere \cite{Hautiere2005,Hautiere2006}, consists of making the computation on a reduced part of the neighbourhood and to pre-calculate some intermediate images of minimum and maximum between the image translated horizontally and vertically in order to compute the contrast. However, this algorithm does not introduce any parallelisation.

In this paper, after a reminder on K\"ohler's method and on parallel computing in Mathematical Morphology \cite{Matheron1967,Serra1982,Soille2003}; we will introduce a parallel algorithm for K\"ohler's method using line translations of the image. We will also propose to compute the contrast on a reduced neighbourhood as in \cite{Hautiere2005}. Eventually, we will compare an implementation of our algorithm, using vectorisation with SIMD instructions \cite{Cockshott2010} and multi-core (i.e. parallel) processing with OpenMP \cite{Chapman2008}, to other implementations of K\"ohler's method.

%%%%%%%%%%%%%%%%%%%%%%%%%%%%%%%%%%%%%%%%%%%%%%%%%%%%%%%%%%%%%%%
%
%		Reminder: Köhler's method
%
%%%%%%%%%%%%%%%%%%%%%%%%%%%%%%%%%%%%%%%%%%%%%%%%%%%%%%%%%%%%%%%
\section{Prerequisites}
\label{sec:pre}

Let us remind K\"ohler's method and the acceleration of Mathematical Morphology operations. An image is a function $f$ defined on a discrete domain $D \subset \Zint^n$ with values in $[0, M[$, $M \in \Real$ and $M=256$ for 8 bits images. We denote $x$ the location of a pixel and $f_x$ its value.
In the sequel, we will use the 4-neighbourhood $N_4$ of pixels. For bi-dimensional images, we can also use the 8 or the 6-neighbourhood \cite{Soille2003} with insignificant differences \cite{Hautiere2006}.

%%%%
%
%		K\"ohler's method
%
%%%%

\subsection{K\"ohler's method}
\label{ssec:Kohler}

Let us remind K\"ohler's method \cite{Kohler1981,Jourlin2012}. Given a grey-level image $f$ and a threshold $t \in [0,M[$, two classes are generated by $t$: $C_0^t(f) = \{ x \in D, f_x \leq t \}$ and $C_1^t(f) = \{ x \in D, f_x > t \}$.
A boundary $B(t)$ is also generated:
\begin{equation}\label{eq:pre:Kohler_boundary}
\begin{array}{@{}ccl@{}}
B(t) &=& \left\{ (x_0,x_1) \in D^2, x_0 \in C_0^t(f), x_1 \in C_1^t(f) \right.\\
 &&\left. \text{ and } x_1 \in N_4(x_0)\right\}.
\end{array}
\end{equation}
For each couple of pixels $(x_0,x_1)$ of $B(t)$, K\"ohler associates a contrast $C_K^t(x_0,x_1)$ defined as:
\begin{equation}\label{eq:pre:Kohler_contrast}
C_K^t(x_0,x_1) = \min\left( f_{x_1}-t , t-f_{x_0} \right)
\end{equation}
which is the minimum of the two steps (in terms of contrast) generated by the threshold $t$ between $f_{x_0}$ and $f_{x_1}$.
The average contrast of the boundary $B(t)$ is defined as:
\begin{equation}\label{eq:pre:Kohler_boundary_contrast}
C_K(B(t)) = \frac{1}{\# B(t)}\times \sum_{(x_0,x_1) \in B(t)} C_K^t(x_0,x_1).
\end{equation}
$\# B(t)$ is the cardinality (number of elements) of $B(t)$ and the summation is made on the couples of pixels $(x_0,x_1)$ belonging to $B(t)$.
This generates a curve of contrasts $C_K(B(t))$ for all the possible thresholds $t \in [0,M[$. The optimal threshold $t_0$ is selected as:
\begin{equation}\label{eq:pre:Kohler_threshold}
C_K(B(t_0)) = \max_{t\in [0,M[} \left( C_K(B(t) \right).
\end{equation}

\begin{figure}[!htb]
\begin{tabular}{@{}c@{ }c@{}}
	\includegraphics[angle=0,width=0.5\columnwidth]{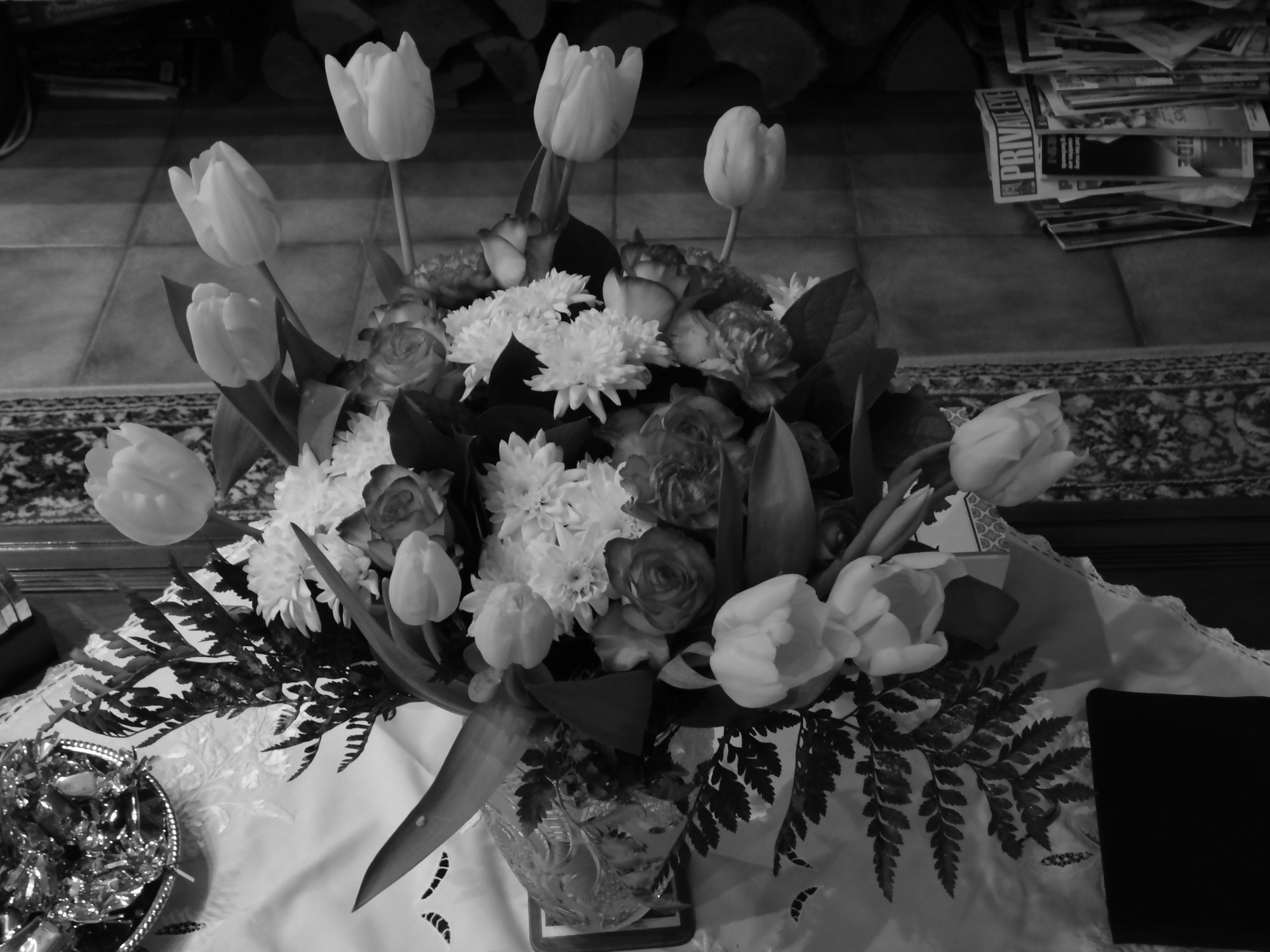}&
	\includegraphics[angle=0,width=0.5\columnwidth]{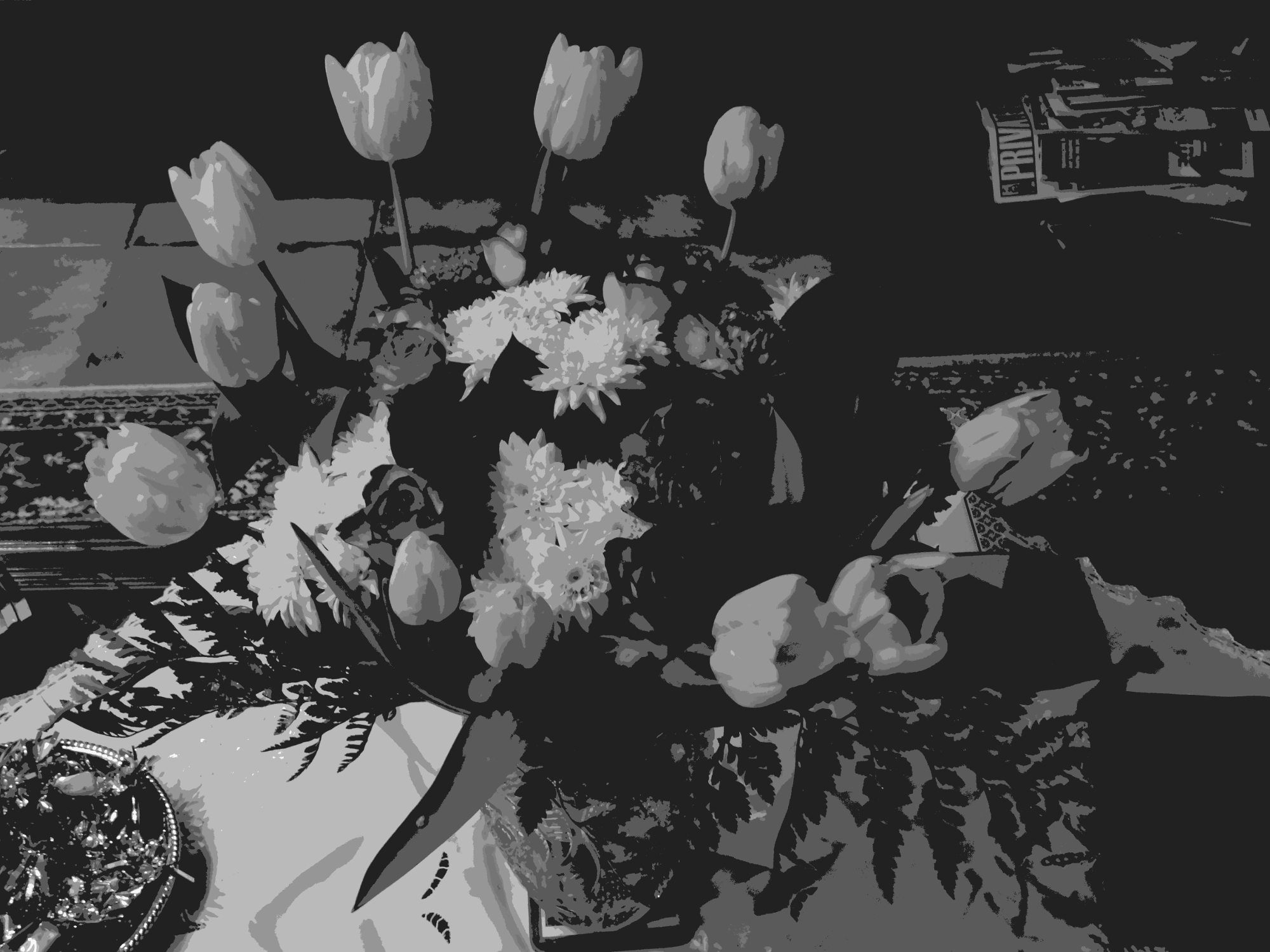}\\
	\footnotesize{(a) Initial image (256 classes)} & \footnotesize{(c) Segmented image (7 classes)}\\
	\multicolumn{2}{c}{\includegraphics[angle=0,width=0.6\columnwidth]{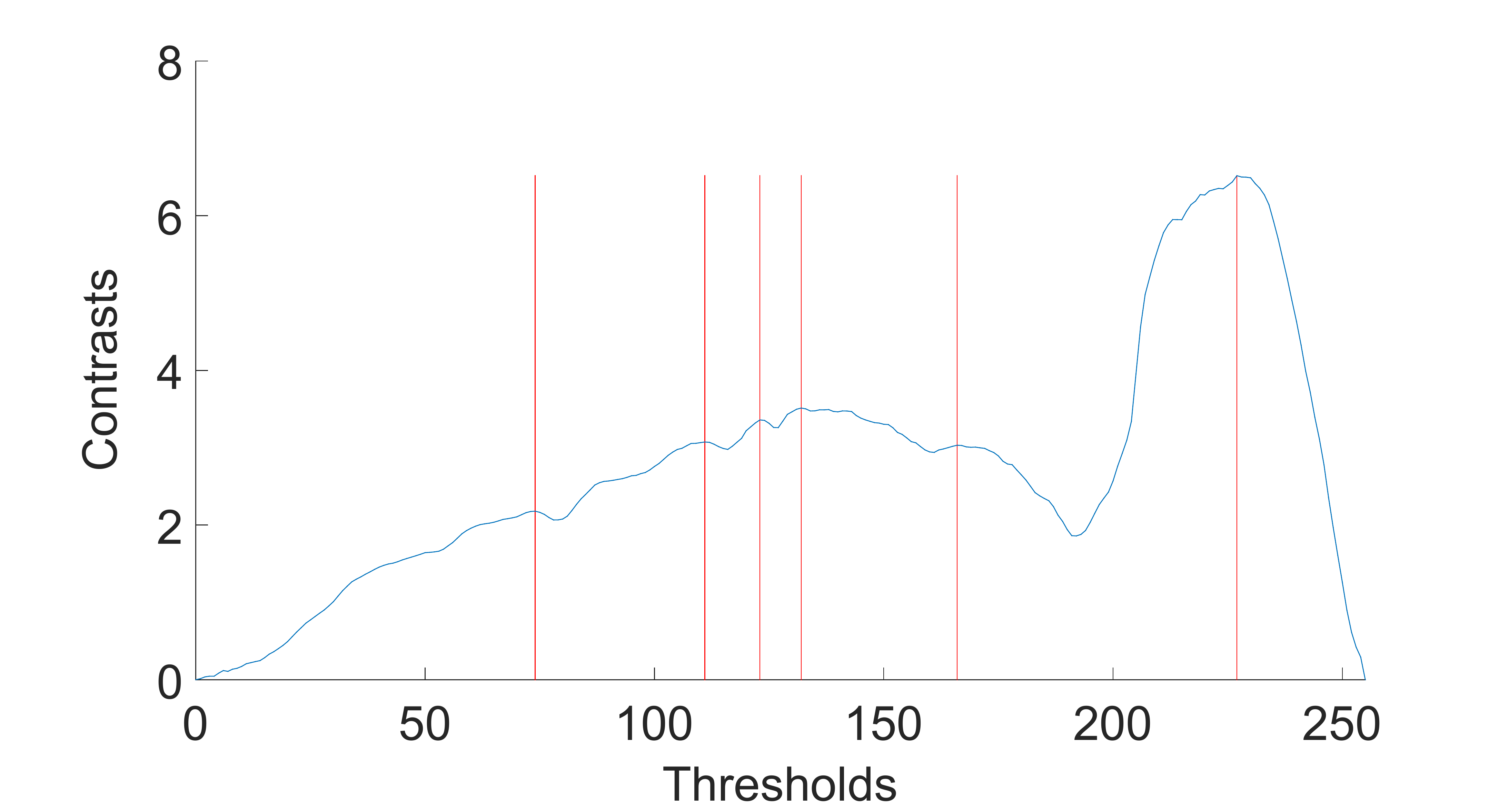}}	\\
	\multicolumn{2}{c}{\footnotesize{(c) Contrast curve} }\\
	\end{tabular}
	\caption{Multiple thresholding by K\"ohler's method of the (a) original image into a (b) segmented image (the class value is the mean values of the class pixels) by (c) the seven thresholds selected on the contrast curve.}
	\label{fig:pre:Kolher_tulips}
\end{figure}

In figure \ref{fig:pre:Kolher_tulips}, we have extracted the 6 most significant thresholds (i.e. the local maxima) from the contrast curve. These multiple thresholds give an efficient simplification (i.e. compression) of the image grey levels: passing from 256 to 7 grey levels. The 7 grey levels corresponds to the mean value of the pixels for each class of the segmentation.

A direct C++ implementation consists of computing for each threshold $t \in[0,M[$, the contrast $C_K^t(B(t))$ of the boundary $B(t)$. It has a duration of 53 s using an image of size $3672 \times 4096$ pixels and a processor Intel\textregistered Core\textsuperscript{TM} i7 CPU 4702HQ, 2.20 GHz, 4 cores, 8 threads. As the algorithm is not parallel, a single thread is used. For real-time applications, or big datasets, a faster algorithm is needed.

%%%%
%
%		Accelerating operations on neighbourhood
%
%%%%
\subsection{Accelerating operations on a neighbourhood}
\label{ssec:accel}

\begin{figure}[htb]
	\begin{tabular}{@{}c|c@{}}
		\includegraphics[angle=0,width=0.5\columnwidth]{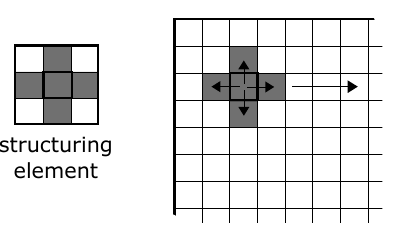}&
		\includegraphics[angle=0,width=0.5\columnwidth]{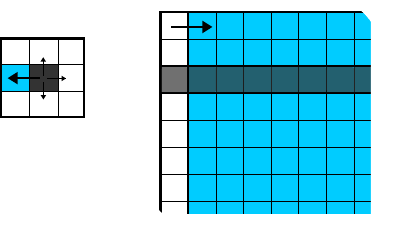}\\
		\footnotesize{(a)} & \footnotesize{(b)}\\
	\end{tabular}
	\caption{Computation of a dilation (a) using a neighbourhood iterator or (b) by line translation.}
	\label{fig:fast:morpho_neigh}
\end{figure}

In Mathematical Morphology, for operations on a neighbourhood some acceleration methods exists. With a symmetric structuring element $A$ (such as the one associated to the 4-neighbours), the morphological dilation corresponds to the Minkowski addition \cite{Minkowski1903,Matheron1967,Serra1982,Soille2003,Najman2013}:
\begin{equation}\label{eq:fast:Minkowski_plus}
	X \oplus A = \bigcup_{a \in A} X_a = \{ x + a : x \in X , a\in A\}.
\end{equation}
$X_a = \{ x + a : x \in X \}$ is the set $X \subset D$ translated by the vector $a$. A direct implementation of a dilation, by computing the union on the neighbourhood of each pixel (fig. \ref{fig:fast:morpho_neigh} (a)), will lead to an algorithm of complexity of $\mathcal{O}(N \times |A|)$ \cite{Geraud2010}. $N$ is the number of pixels in the image and $|A|$ is the cardinality of $|A|$. However, the Minkowksi addition (i.e. the dilation) has the property to distribute the union \cite{Hadwiger1957,Serra1982}:	$X \oplus (A \cup A') = (X \oplus A) \cup (X \oplus A')$. This property has important technological consequences as a dilation can be computed elements by elements of the structuring element $A$, before combining the intermediate results by union. Therefore,
\begin{equation}\label{eq:pre:dil_line}
\begin{array}{ccl}
	X \oplus A &=& \{ x \in D \> | \> \exists a \in A, x - a \in X \}\\
	&=& \bigcup_{a \in A} \{ x \in D \> | \> x - a \in X \}.
	\end{array}
\end{equation}
An implementation by translating all pixel $x$ by the vectors $a \in A$ and checking if they belong to the set $X$  will have a complexity of $\mathcal{O}(|X \oplus A|)$ \cite{Geraud2010}. As images are stored in computer memory as unidimensional arrays, an efficient implementation \cite{Faessel_smil2013} consists of translating the lines instead of the pixels (fig. \ref{fig:fast:morpho_neigh} (b)). 
As the operations in each line are independent from the other lines (eq. \ref{eq:pre:dil_line}), a parallel programming method is used, namely OpenMP \cite{Chapman2008}, which is designed for multi-processor/core, shared memory computers. Each thread is computed by a single core and all cores share a common memory. In addition, vectorised data are used with SIMD instructions \cite{Cockshott2010} in each thread. Vector operations using SIMD instructions allow multiple data to be processed with a single instruction of the processor while scalar operations use one instruction to process each individual data. The size of the SIMD registers being of 128 bits, 16 operations on integers of 8 bits are performed at the same time instead of a single one. In recent compilers, the vectorisation is performed automatically after activation of the right option. In \cite{Faessel_smil2013}, a dilation with a square structuring element of size 3 pixels in a 8 bit image (of size $1024 \times 1024$ pixels) is accelerated by a factor 136 with a neighbourhood implementation by comparison to a line implementation with parallelisation and vectorisation (45 ms versus 0.33 ms, Intel\textregistered Core\textsuperscript{TM} i3 CPU M330, 2.13 GHz, 2 cores, 4 threads). Let us introduce an efficient method to compute K\"ohler's method.

%%%%%%%%%%%%%%%%%%%%%%%%%%%%%%%%%%%%%%%%%%%%%%%%%%%%%%%%%%%%%%%
%
%		A fast algorithm of K\"ohler's method
%
%%%%%%%%%%%%%%%%%%%%%%%%%%%%%%%%%%%%%%%%%%%%%%%%%%%%%%%%%%%%%%%
\section{A fast algorithm for K\"ohler's method}
\label{sec:fast}

%Let us present the acceleration method for K\"ohler's contrast, the reduction of the neighbourhood size and a fast algorithm of K\"ohler's method.

%--------------------------------------------------------------
%		Accelerating the computation of K\"ohler's contrast
%
%--------------------------------------------------------------
\subsection{Accelerating the computation of K\"ohler's contrast}
\label{ssec:accel_Kohler}

A direct implementation of K\"ohler's approach (section \ref{ssec:Kohler}) is not designed for parallel and vector processing. 
K\"olher's contrast (eq. \ref{eq:pre:Kohler_contrast}) is summed on boundaries $B(t)$ which are computed by the set difference between a morphological dilation of the set $C_1^t(f)$ and this same set: $\{x_0 \in D, (x_0,x_1) \in B(t)\} = (C_1^t(f) \oplus A) \setminus C_1^t(f)$. The structuring element $A$ corresponds to the 4-neighbours. The direct implementation has a complexity of $\mathcal{O}(N^2)$ (i.e. the complexity of a dilation, $\mathcal{O}(N)$, multiplied by the complexity of scanning the pixel pairs of the boundary $B(t)$, $\mathcal{O}(M \times \#B(t)) = \mathcal{O}(N)$). A direct acceleration would consist of computing the boundary with an accelerated morphological dilation, as presented above. However, such approach does not reduce the complexity of the algorithm. For this purpose, we propose first to perform the translation of the image lines, which is suited for parallel processing, and then to compute the contribution to the contrast of each pixel pairs, for each threshold. The complexity decreases to $\mathcal{O}(NM)$, i.e. the product of the number of pixels by the number of grey levels. The algorithm \ref{alg:fast_Kohler} presents our approach.

\begin{algorithm}[!htb]
\caption{Fast algorithm for K\"ohler's method} \label{alg:fast_Kohler}
  \begin{algorithmic}[1]
	%\algsetup{indent = 2em}
	\State{$I$, $J$}					 																	\Comment{number of lines and columns}
	\State{$inLine(i , :)$} 																		\Comment{Line $i$ of the input image} 
	\State{$a$} 																								\Comment{$a=(a.x,a.y)$ translation of the neighb. $A=N_4^{*}$}
	\State{$curLine$, $nLine_1$ ... $nLine_a$} 					      	\Comment{arrays of length $J$} 
	\State{$C$, $card$} 																				\Comment{arrays of length $M$ initially set to zero} 
    \ForAll{ $i \in [1 , I]$}																	\label{alg:fast_Kohler:begin_parallel}
			\State{$curLine \leftarrow inLine(i , :)$ }
			\ForAll{ $a \in A$ }																				\label{alg:fast_Kohler:line_multicore}
				\State{$nLine_{a} \leftarrow translate( inLine(i+a.y , :) , a.x$ )} \label{alg:fast_Kohler:translate_line}%\Comment{see alg. \ref{alg:app:translate_line}}
				\ForAll{ $j \in [1 , J]$}															\label{alg:fast_Kohler:autov_1}
					\State $mini \leftarrow \min( curLine(j) , nLine_{a}(j) )$ \label{alg:fast_Kohler:mini}
					\State $maxi \leftarrow \max( curLine(j) , nLine_{a}(j) )$ \label{alg:fast_Kohler:maxi}
					%\If{ $nLine_{b}(j) \leq t$}
						\ForAll{ $t \in [mini , maxi-1]$ }
							\State{ $C(t) \leftarrow C(t) + \min( maxi - t , t - mini )$} \label{alg:fast_Kohler:contrast}
							\State{ $card(t) \leftarrow card(t) + 1$ }										\label{alg:fast_Kohler:counter}
						\EndFor
					%\EndIf
				\EndFor
			\EndFor
		\EndFor 																									\label{alg:fast_Kohler:end_parallel}
		
		\ForAll{ $t \in [0 , M[$ }																\label{alg:fast_Kohler:norm_contrast}
			\State{$C(t) \leftarrow C(t) / card(t)$}
		\EndFor
  \end{algorithmic}
\end{algorithm}

%--------------------------------------------------------------
%		Reduction of the neighbourhood size
%
%--------------------------------------------------------------
\subsection{Reduction of the neighbourhood size}
\label{ssec:reduc_neigh}

\begin{figure}[!htb]
	\begin{center}
	\begin{tabular}{cc}
	\includegraphics[angle=0,width=0.2\columnwidth]{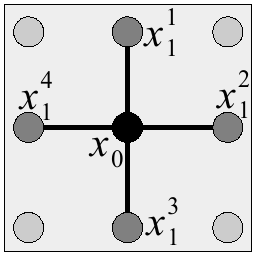}&
	\includegraphics[angle=0,width=0.2\columnwidth]{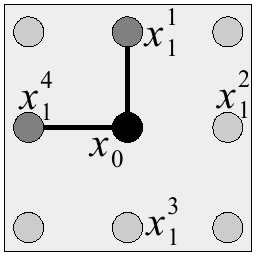}\\
	\footnotesize{(a) $N_4$} & \footnotesize{(b) $N_4^*$}
	\end{tabular}
	\end{center}
	\caption{(a) The 4-neighbourhood $N_4$ and its (b) ``half'' $N_4^*$}
	\label{fig:reduc:neigh}
\end{figure}

In order to gain an additional factor 2, let us show that it is the same to make the computation on a ``half'' neighbourhood $N_4*$ as on the 4-neighbourhood $N_4$ (fig. \ref{fig:reduc:neigh}). The idea has been introduced in \cite{Hautiere2005,Hautiere2006}. Here, the equality of the two approaches is demonstrated. 

When scanning the boundary $B(t)$ with the 4-neighbourhood $N_4$, only the pixels $x_1$, such as $f_{x_0} < f_{x_1}$, are contributing to the boundary contrast. 
However, both pixels $x_0$ and $x_1$ are neighbours: $x_0 \in N_4(x_1) \Leftrightarrow x_1 \in N_4(x_0)$. Therefore,  
both (ordered) pairs $(x_0,x_1)$ and $(x_1,x_0)$ are scanned and only one pair is contributing to the contrast between the (unordered) set of points $\{x_0,x_1\}$
\begin{equation} \label{eq:dem_cont2}
\begin{array}{@{}c@{ }c@{ }l@{}}
	\Gamma_{K}^t(\{x_0,x_1\}_{N_4}) &=& 
	\left\{
	\begin{array}{@{}l@{}}
	\min\left( f_{x_0}-t , t-f_{x_1} \right), \text{if } f_{x_1} \leq t < f_{x_0}\\
	\min\left( f_{x_1}-t , t-f_{x_0} \right), \text{else }f_{x_0} \leq t < f_{x_1}
	\end{array}	 \right.\\
	&=& \min\left( |f_{x_0}-t| , |t-f_{x_1}| \right).\\
\end{array}
\end{equation}
Let us remove the order condition between the grey levels, $f_{x_0} \leq t < f_{x_1}$ and define the absolute pair contrast $\overline{C}_{K}^t(x_0,x_1) =  \min\left( |f_{x_0}-t| ,\right.$ $\left.|t-f_{x_1}| \right)$. When scanning both pairs without the grey level order, the contrast between the set of points $\{x_0,x_1\}$, is:
\begin{equation} \label{eq:dem_cont3}
\begin{array}{@{}c@{ }c@{ }l@{}}
	\overline{\Gamma}_{K}^t(\{x_0,x_1\}_{N_4}) &=& \overline{C}_{K}^t(x_0,x_1) + \overline{C}_{K}^t(x_1,x_0)\\
	&=& 2 \min\left( |f_{x_0}-t| , |t-f_{x_1}| \right)\\
	&=& 2 \Gamma_{K}^t(\{x_0,x_1\}_{N_4}), \quad \text{(eq. \ref{eq:dem_cont2})}\\
\end{array}.
\end{equation}
Using the ``half''-neighbourhood $N_4^*$ allows to scan only one pair of pixels. Therefore, we obtain our result:
\begin{equation} \label{eq:cont_equal}
	\overline{\Gamma}_{K}^t(\{x_0,x_1\}_{N_4^*}) = \Gamma_{K}^t(\{x_0,x_1\}_{N_4}).
\end{equation}

%--------------------------------------------------------------
%		Implementation: parallelisation and vectorisation
%
%--------------------------------------------------------------
\subsection{Implementation: parallelisation and vectorisation}
\label{ssec:impl}

In order to make parallel the algorithm, the computation of the contrast can be performed independently line by line. For each parallel thread $k$, two arrays of length $M$, $C_k$ (contrast) and $Card_k$ (counter), need to be created at the beginning of the parallel process (line \ref{alg:fast_Kohler:begin_parallel}, alg. \ref{alg:fast_Kohler}). At the end of the parallel process (line \ref{alg:fast_Kohler:end_parallel}), they are grouped by summation in two arrays $C$ (contrast) and $Card$ (counter). The parallel programming language used is OpenMP in C++. Instead of being performed between single numbers, several operations can be performed using arrays, allowing the vectorisation of the data. The following operations are vectorised and processed using SIMD instructions: $i)$ the line translation (line \ref{alg:fast_Kohler:translate_line}), $ii)$ the computation of the minimum (line \ref{alg:fast_Kohler:mini}) and the maximum (line \ref{alg:fast_Kohler:maxi}) between the arrays $curLine$ and $nLine_a$, $iii)$ the computation of the contrast $C_k$ (line \ref{alg:fast_Kohler:contrast}) and of the counter $card_k$ (line \ref{alg:fast_Kohler:counter}) and $iv)$ the normalisation of the contrast (line \ref{alg:fast_Kohler:norm_contrast}).

%%%%%%%%%%%%%%%%%%%%%%%%%%%%%%%%%%%%%%%%%%%%%%%%%%%%%%%%%%%%%%%
%
%		Results
%
%%%%%%%%%%%%%%%%%%%%%%%%%%%%%%%%%%%%%%%%%%%%%%%%%%%%%%%%%%%%%%%
\section{Results}
\label{Results}

We now compare the duration of the direct implementation to the fast algorithm. We have used a processor Intel\textregistered Core\textsuperscript{TM} i7 CPU 4702HQ, 2.20 GHz, 4 cores, 8 threads with 16Gb RAM. Using the image ``Tulips'' (fig. \ref{fig:pre:Kolher_tulips}) with a current camera  resolution of $3672 \times 4896$ pixels and the fast algorithm with parallelisation, the computation of K\"ohler's method is made in 0.13 s (tab. \ref{tab:res:Duration}) instead of 53 s, with a gain factor of 405. With images of former resolution ($512 \times 512$ pixels), such as Lenna image; the direct method takes 0.69s and the fast method 0.005s with a gain factor of 126. Therefore, the necessity of using a faster algorithm instead of direct implementation becomes essential to process images with current resolution. Other experiments have confirmed this result.

\begin{table}[!htb]
\begin{center}
	\begin{tabular}{lcc}
	\hline
	Name 					&  Lenna & Tulips\\
	\hline
	Size (pixels)	& $512 \times 512$ 	& $3672 \times 4896$\\
	Direct (D) 		& 6.90e-01 s				& 5.29e+01 s\\
	Fast (B)			&	1.62e-02 s  			& 4.86e-01 s\\
	Fast (A)	 		& 5.46e-03 s 				& 1.30e-01 s\\
	Gain (B vs. D) 	& 42 								& 109\\
	Gain (A vs. D) 	& 126								& 405\\
	\hline
	\end{tabular}
	\end{center}
	\caption{Comparison of the duration of the different implementations for images of different sizes: direct (D), fast without parallelisation (B) and fast with parallelisation (A). The gain factors are computed between the implementations B and D and between the implementations A and D.}
	\label{tab:res:Duration}
\end{table}

\begin{figure}[!htb]
	\begin{center}
	\begin{tabular}{@{}c@{ }c@{}}
	\includegraphics[angle=0,width=0.5\columnwidth]{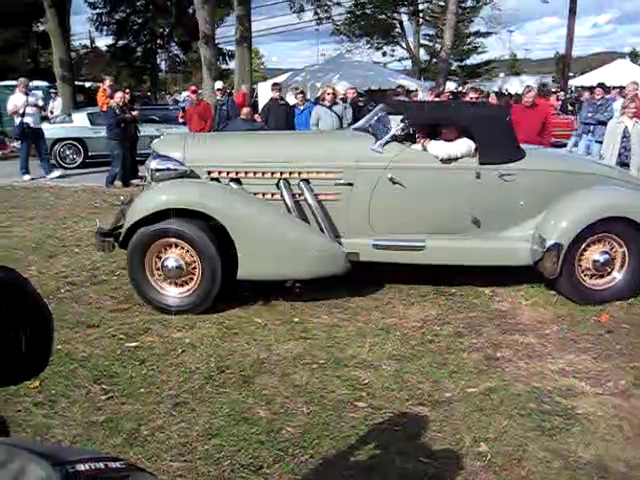}&
	\includegraphics[angle=0,width=0.5\columnwidth]{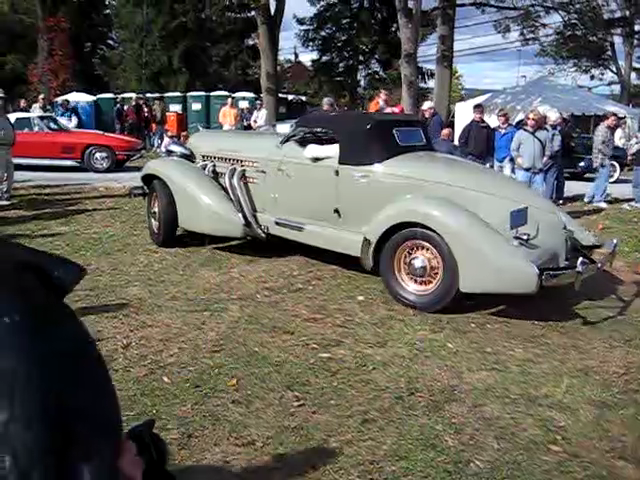}\\
	\footnotesize{(a) frame 101} & \footnotesize{(b) frame 251}\\
	\includegraphics[angle=0,width=0.5\columnwidth]{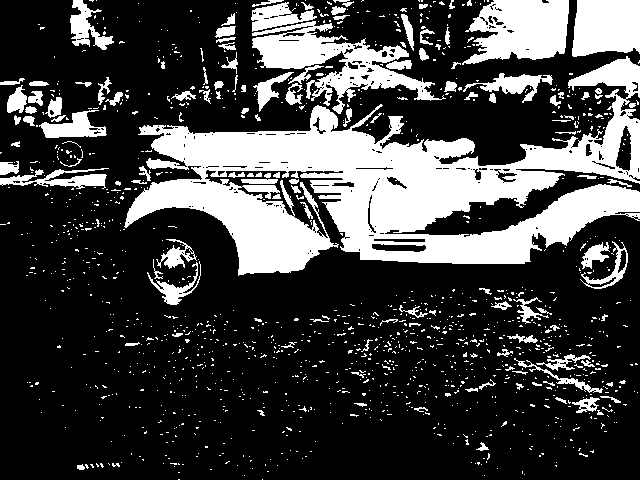}&
	\includegraphics[angle=0,width=0.5\columnwidth]{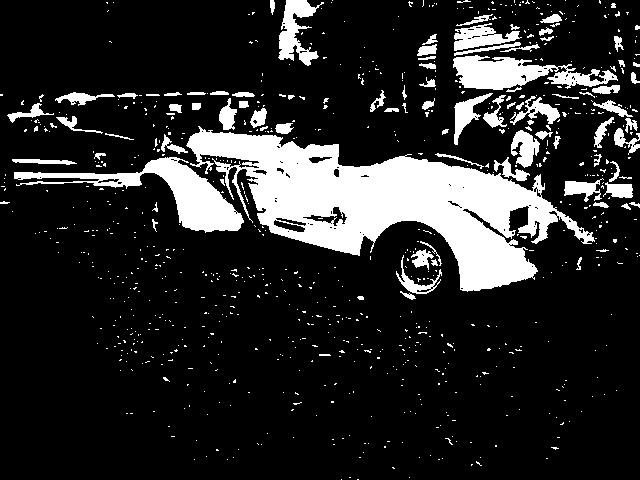}\\
	\footnotesize{(c) segmented frame 101} & \footnotesize{(c) segmented frame 251}\\	
	\end{tabular}
	\end{center}
	\caption{(a), (b) Two frames of a video from the dataset YFCC100M and (c), (d) their segmentations in two classes by K\"ohler's method.}
	\label{fig:res:frames_video}
\end{figure}

Let us try K\"ohler's method with a video of a car from the dataset YFCC100M (Yahoo Flickr Creative Commons 100M) \cite{Thomee2016,YFCC100M_video_car2009}. In figure \ref{fig:res:frames_video}, two frames and their segmentations in two classes are shown. The direct implementation segments the video at a rate of 1 frame per second while the fast implementation (with parallelisation) processes 97 frames per second, which is faster than the 25 frames/s of the video (tab. \ref{tab:res:Duration_video}). Therefore, the fast algorithm for K\"ohler's method is suited for real time video processing. 

\begin{table}[!htb]
\begin{center}
	\begin{tabular}{lcc}
	\hline
	Name 					&  YFCC100M (car) & \\
	\hline
	Size (pixels)	& $502 \times 480$ 	& \\
	Number of frames & 640 &\\
	Direct (D) 		& 0.99  frames/s			& \\
	Fast (A)	 		& 96.96 frames/s 		& \\
	Gain (A vs. D) 	& 98								& \\
	\hline
	\end{tabular}
	\end{center}
	\caption{Frame per seconds segmented by K\"ohler's method with different implementations applied on a video: direct (D), and fast with parallelisation (A). The gain factors between the implementations A and D have been computed.}
	\label{tab:res:Duration_video}
\end{table}

%Several other experiments were performed in images and in video at different resolutions. All have shown similar gain factors to the reported experiments . 

\section{Conclusion and perspectives}
\label{sec:concl}

A faster algorithm for K\"ohler's thresholding has been introduced with a lower complexity, $\mathcal{O}(NM)$, than the direct approach, $\mathcal{O}(N^2)$. It is designed to benefit from the capacities of processors: multi-core processing with OpenMP and vector processing using SIMD instructions. Results show that with an image of 18 million pixels the duration is reduced by a factor 405 (from 53 s to 0.13 s) and that a video can be processed at a rate of 97 frames/s instead of 1 frame/s. Importantly, this algorithm is suited for applications requiring real-time or fast processing: video, industrial, large databases, etc. Its practical interest is to be combined with previous transforms: a low-pass filter, a mathematical morphology transform \cite{Serra1982,Serra1988,Geraud2010,Noyel2010} or a map of colour distances \cite{Noyel2015,Noyel2016}. In future works, the influence on the method of different contrasts will be presented (already studied), such as the contrasts defined in the Logarithmic Image Processing framework \cite{Jourlin2012,Jourlin2016_chap3}.

% References should be produced using the bibtex program from suitable
% BiBTeX files (here: strings, refs, manuals). The IEEEbib.bst bibliography
% style file from IEEE produces unsorted bibliography list.
% -------------------------------------------------------------------------
\bibliographystyle{IEEEbib}
\bibliography{refs}

\end{document}